\pdfoutput=1

\documentclass[11pt]{article}

\usepackage[preprint]{acl}

\usepackage{times}
\usepackage{latexsym}
\usepackage{booktabs} 

\usepackage[T1]{fontenc}

\usepackage[utf8]{inputenc}

\usepackage{microtype}

\usepackage{inconsolata}

\usepackage{graphicx}

\title{\includegraphics[width=0.7cm,keepaspectratio]{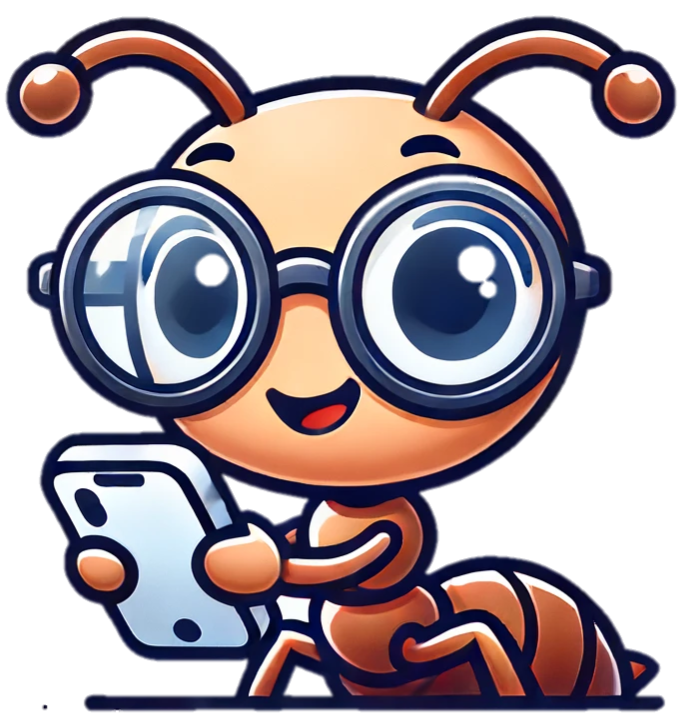}E-ANT: A Large-Scale Dataset for Efficient Automatic GUI NavigaTion}

\author{
 \textbf{Ke Wang \textsuperscript{1}\thanks{Equal contribution}},
 \textbf{Tianyu Xia \textsuperscript{1}\footnotemark[1] },
 \textbf{Zhangxuan Gu\textsuperscript{1}},
 \textbf{Yi Zhao\textsuperscript{2}},
\\
 \textbf{Shuheng Shen\textsuperscript{1}},
 \textbf{Changhua Meng\textsuperscript{1}},
 \textbf{Weiqiang Wang \textsuperscript{1}},
 \textbf{Ke Xu \textsuperscript{2}},
\\
 \textsuperscript{1}Ant Group,
 \textsuperscript{2}Tsinghua University
\\
\texttt{ke.wang\_kay@foxmail.com, xty368914@antgroup.com}
}

\begin{document}
\maketitle
\begin{abstract}
 Online GUI navigation on mobile devices has driven a lot of attention recent years since it contributes to many real-world applications. With the rapid development of large language models (LLM), multimodal large language models (MLLM) have tremendous potential on this task. However, existing MLLMs need high quality data to improve its abilities of making the correct navigation decisions according to the human user inputs. In this paper, we developed a novel and highly valuable dataset, named \textbf{E-ANT}, as the first Chinese GUI navigation dataset that contains real human behaviour and high quality screenshots with annotations, containing more than 40,000+ real human traces over 20000+ different tinyAPPs and URLs. Furthermore, we evaluate various powerful MLLMs on E-ANT and show their experiments results with sufficient ablations. We believe that our proposed dataset will be beneficial for both the evaluation and development of GUI navigation and LLM/MLLM decision-making capabilities.
\end{abstract}
 
\section{Introduction}
\label{sec:intro}
The integration of natural language and voice commands for automating tasks on mobile devices is a pivotal topic within human-computer interaction and intelligent agent design. This research holds immense value for individuals with physical disabilities or those engaged in activities like driving, where hands-free operation is essential. An efficient mobile device automatic control system should continuously comprehend the active screen, make informed decisions, and execute the necessary actions to fulfill objectives articulated through natural language, such as automatically opening vehicle navigation or ordering food while driving.
\begin{figure}
    \centering
    \includegraphics[width=0.30\columnwidth]{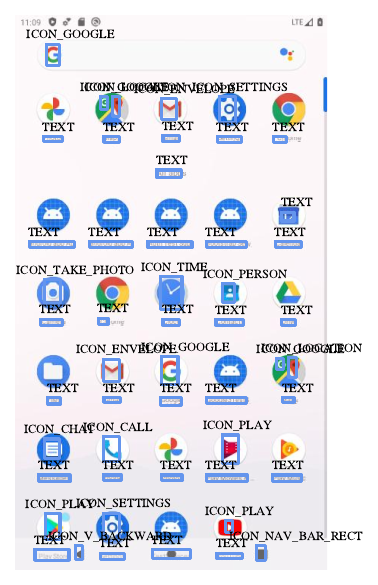}
    \includegraphics[width=0.29\columnwidth]{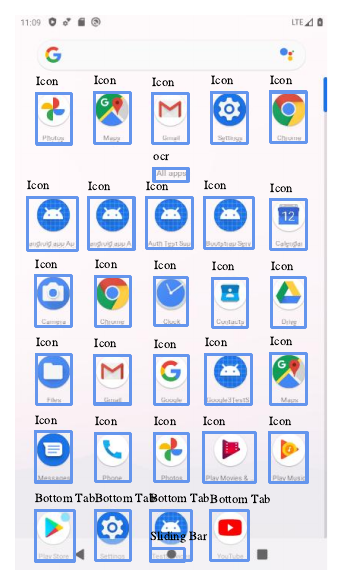}
    \includegraphics[width=0.32\columnwidth]{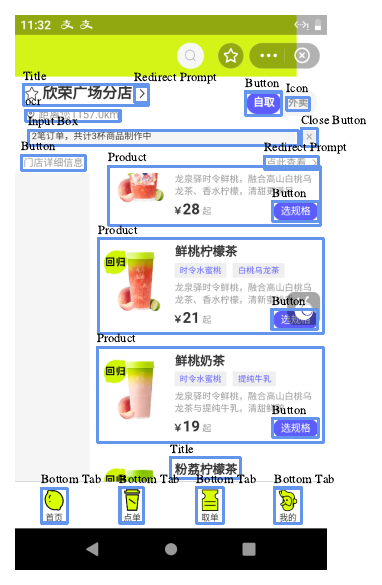}
    \caption{Left: Annotations of AitW. Middle: Our page analysis results on AitW data. Right: Our page analysis results on our dataset. Note: Our method accurately identifies the "Select Specifications" button, critical for order placements.}
    \label{fig:layoutcase}
    \vspace{-4mm}
\end{figure}

Existing research approaches are from either a software engineering~\cite{adb,xcode,li2017sugilite,azim2016ulink,li2017programming} or navigation algorithms~\cite{li2020mapping,bai2021uibert,hong2023cogagent,rawles2024androidinthewild,wang2023enabling,wen2023droidbot,wen2023empowering,yan2023gpt,zhang2023reinforced}. The former methods explore the automation of instruction execution or the abstraction of various APIs, while the latter ones are primarily concerned with translating natural language commands into system-comprehensible instructions (such as clicks or slides). 
Nevertheless, they both need the model has powerful decision-making capability given human inputs.
In this case, GUI navigation benchmark is an essential aspect of evaluating the decision-making capabilities of both large language models(LLMs)~\cite{achiam2023gpt,zheng2024judging,touvron2023llama,qwen,baichuan2023baichuan2,jiang2023mistral,falcon40b} and multi-modal large language models(MLLMs)~\cite{liu2024visual,liu2023improved,zhu2023minigpt,chen2023minigptv2,li2022blip,li2023blip2,wang2023cogvlm,hong2023cogagent}. It holds significant importance in assessing their performance as agents.
As a result, Android in the Wild (AitW)~\cite{rawles2024androidinthewild} is a data benchmark recently introduced by Google to assess the efficiency of UI navigation algorithms while performing daily tasks on native Android systems. This benchmark effectively fills the void in data benchmarks for evaluating UI interaction within Android systems.

Despite the proliferation of Chinese mobile applications, surpassing 2.61 million~\cite{miit_policy_2023}, the majority of existing GUI navigation datasets primarily cater to English, leading to a clear lack of comprehensive datesets for Chinese GUI navigation. Additionally, datasets like AitW are narrowly focused on GUI navigation within the native Android operating system and its inherent applications, making their applicability to third-party apps from various developers limited. Moreover, the quality of annotation regarding GUI element positions in these datasets is poor, with some inaccuracies(figure~\ref{fig:layoutcase}) and wrong labels~\cite{hong2023cogagent}, which can impair the decision-making precision in downstream GUI navigation activities.

\begin{figure*}[tb]
  \centering
  \includegraphics[width=0.98\textwidth]{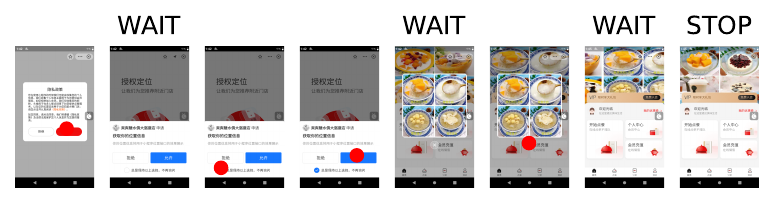}
  \caption{Sample Trajectory in Our Dataset: Navigating to a Merchant's Homepage in a Tiny-App (Red Dot Indicates Click Location in the Current State).
  }
  \label{fig:label-num-trace}
  \vspace{-4mm}
\end{figure*}

In this paper, we focus on navigating the Chinese GUI in third-party applications created by various developers. We primarily gather our data from tiny-apps, which are lightweight and simple to develop mobile applications. Currently, there are over 4.6 million active tiny-apps.
We developed a Large-Scale Dataset for \textbf{E}fficient \textbf{A}utomatic GUI \textbf{N}aviga\textbf{T}ion (E-ANT), 
consisting of over 40,000 user operation trajectories. This dataset covers a wide range of navigation intentions and includes various tiny apps. Comparatively, interacting with the Android native system is different as tiny apps are typically created by numerous third-party developers, each with their own design logic and art styles. This diversity poses unique challenges in navigating through tiny apps.
We provide a clear and comprehensive understanding of each trajectory. This includes an intention described in natural language, a series of consecutive page screenshots ranging from several to dozens, and the corresponding actions performed on each page, such as clicks and slides in coordinate dimensions. Additionally, for each page screenshot, we offer detailed information about the page elements captured, including their type (such as button, icon, OCR, etc.), coordinates, and the text contained within each element.

To gain a precise understanding and evaluation of the GUI navigation performances of current mainstream LLMs/MLLMs, we conducted extensive benchmark tests on E-ANT. Specifically, we evaluate the GUI navigation level of the current mainstream models under the following strategies.
\textbf{(1)Zero-shot inference}. Directly use the existing pre-trained model to test on the test set. 
\textbf{(3)Fine-tuning}. Use a part of the samples as a training set to fine-tune the model before inference.
\textbf{(4)Fine-tuning with data augmentation}. This is our recommended method of fine-tuning the UI navigation model. It does not directly use coordinate positions as labels, but allows them to make decisions step by step in a chain. We will introduce this method in detail later.

We summarize our contributions as follows: 
\begin{itemize}
 \item We gather and publish the first large-scale Chinese dataset for GUI navigation, collected from diverse tiny apps. It will make foreseeable contributions to both the multimodal and the automatic GUI navigation community.
 \item We analyze the characteristics of our dataset and provide a recommended fine-tuning methods for GUI navigation data.
 \item We evaluate in detail the performance of current mainstream LLMs/MLLMs on this dataset under different inference methods.
\end{itemize}
\vspace{-2mm}
\section{Related Work}

\subsection{UI Navigation and Automation Execution}

Previously, three primary methods existed for incorporating automated UI navigation on mobile devices. These options included smart assistants developed by mobile phone manufacturers such as Siri, as well as macro recording tools~\cite{rodrigues2015breaking,rodrigues2014swat,li2021learning} and Programming by Demonstration (PBD) systems~\cite{cypher1993watch,lieberman2001your,guibert2004example,maues2013keep,li2017sugilite}.
They can all translate user intentions into low-level operations and automate execution.
The smart assistant
is restricted to calling only the built-in applications on mobile phones and a few select external applications that are in collaboration with mobile phone manufacturers. This limitation significantly constrains its range of application scenarios.
The macro recording tool's~\cite{rodrigues2015breaking,rodrigues2014swat,li2021learning} capabilities are limited to playing back user-recorded operations. It lacks the ability to handle tasks with altered parameters or customized actions.
The PBD system not only supports the automatic generation of execution scripts through user demonstrations but also provides corresponding interfaces for users to edit scripts~\cite{cypher1993watch,lieberman2001your,guibert2004example,maues2013keep,li2017sugilite}. However, despite its usefulness, the system still has a learning curve, and the scripts are not easily applicable across various applications with similar functions.

\subsection{UI Navigation with LLMs/MLLMs}
The increasing popularity of large-scale language models and multi-modal language models, including GPT~\cite{achiam2023gpt}, LLaMA~\cite{touvron2023llama}, BaiChuan~\cite{baichuan2023baichuan2}, LLaVA~\cite{liu2024visual,liu2023improved}, MiniGPT4-V~\cite{zhu2023minigpt,chen2023minigptv2}, and others, has led to a growing interest among researchers in utilizing these models as intelligent agents for automating UI navigation~\cite{wang2023enabling,kim2024language,wen2023droidbot,zhang2023responsible,wen2023empowering,lee2023explore,yan2023gpt,zhan2023you,hong2023cogagent,yang2023appagent}. 

Among them, some works~\cite{wang2023enabling, kim2024language, yan2023gpt, wen2023empowering} utilize trained LLMs/MLLMs like GPT. They achieve automatic navigation on mobile devices by prompt and incorporating the knowledge of UI navigation domain in pre-trained LLMs/MLLMs. 
Furthermore, Li et al. ~\cite{li2023zero}introduced structured self-reflection into the UI navigation agent to improve its planning capabilities, while Zhang et al.~\cite{zhan2023you} used Chain-of-Action prompts to improve the performance of multi-modal agents on UI navigation tasks.
In addition, Zhang et al.~\cite{zhan2023you} and Hong et al.~\cite{hong2023cogagent} based on pre-trained LLMs/MLLMs, fine-tuned instructions for the content in the UI navigation field to enhance the accuracy of the model's navigation decisions.

\subsection{UI Navigation Benchmark}

There are currently some studies focusing on evaluation data collection and benchmark construction in the field of UI navigation~\cite{shi2017world,liu2018reinforcement,yao2022webshop,rawles2024androidinthewild,bai2021uibert,deka2017rico}. 
MiniWob~\cite{shi2017world} and MiniWob++~\cite{liu2018reinforcement} are established benchmarks in the field of computer UI navigation research. It requires agents to perform specific tasks in the computer environment they build through instructions such as clicks and inputs. MiniWob++ goes a step further by providing programmatically defined rewards for each decision made during execution.
In addition, WebShop~
\cite{yao2022webshop} provides the UI navigation community with a simulated e-commerce environment. In this environment, agents need to navigate multiple types of web pages and issue different actions to find and purchase products based on instructions.
These environments, datasets, and benchmarks focus primarily on navigation and decision-making on web pages~\cite{shi2017world,liu2018reinforcement,yao2022webshop}.
For mobile phones, UIbert~\cite{bai2021uibert} and RICO~\cite{deka2017rico} provide practical page understanding benchmarks that can effectively evaluate the target detection and recognition capabilities of models or agents on the page. However, they lack the intentions and operational actions actually performed on the page and cannot evaluate the agent's Navigation and decision-making skills.
The AitW dataset~\cite{rawles2024androidinthewild} fills this gap by providing an extensive collection of over 6 million images and corresponding actions performed on the Android operating system. However, its focus is mostly limited to first-party applications, such as settings, clock, and Google Maps, with minimal support for third-party applications with different design styles. 
This limits the comprehensiveness of the evaluation capabilities of this dataset and the generalizability of models trained on this dataset.




\begin{figure*}[tb]
  \centering
  \includegraphics[width=0.98\textwidth]{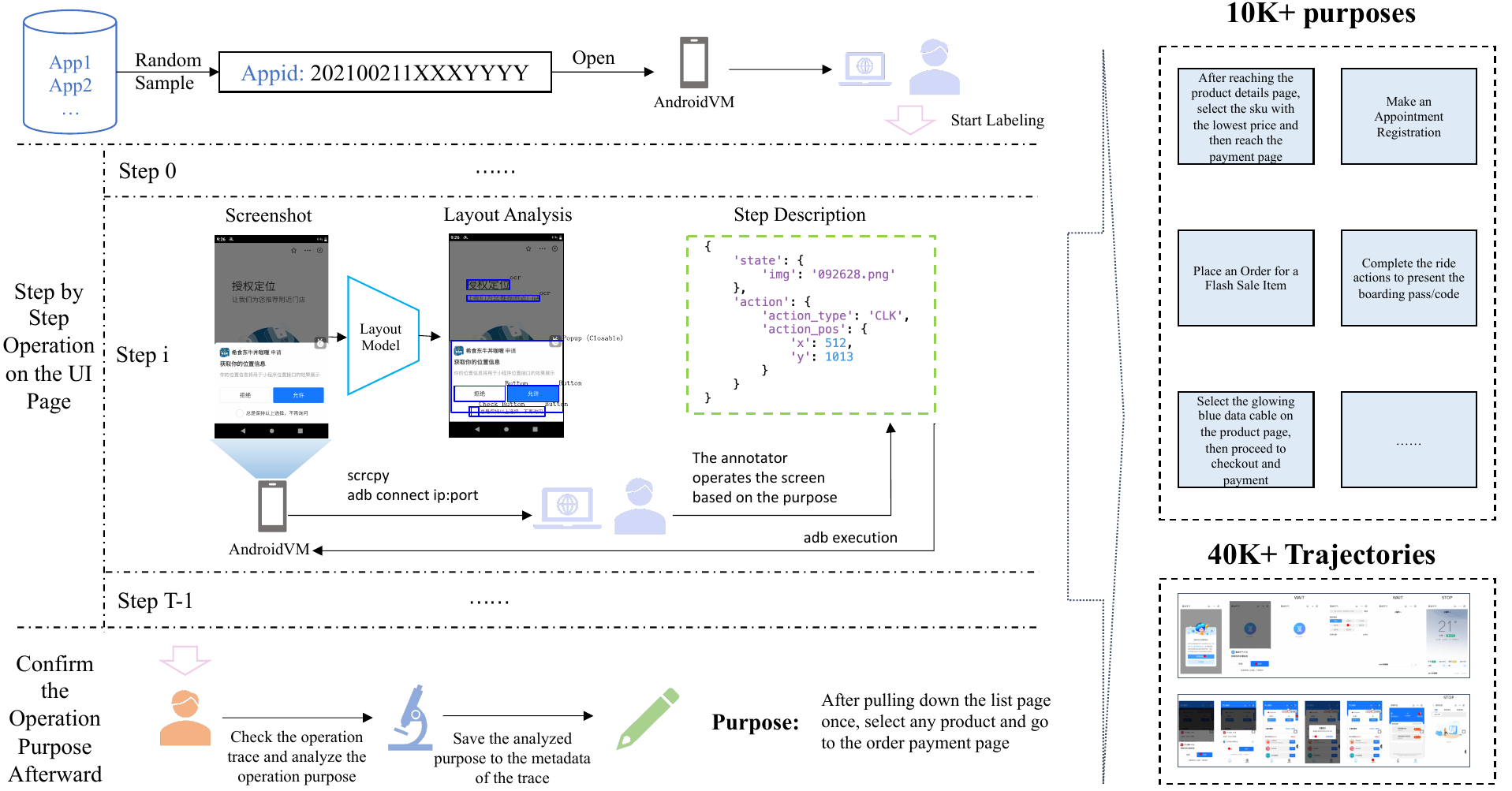}
  \caption{Two steps of sampling from a trajectory: Data overview and data production pipeline for each step in the dataset.
  }
  \label{fig:screenshot}
  \vspace{-4mm}
\end{figure*}

\section{E-Ant TinyAPP Dataset}
\subsection{GUI Navigation Task}

Generally, navigating a GUI can be seen as a series of decision-making tasks when interacting with a webpage or app.

\textbf{Decision-making tasks on UI pages.} For a web page denoted as $S_t$, it hosts a range of interactive controls. We identify the collection of all possible interactions within these elements as the action set $A_t$. Upon selection and execution of an action $a_{t}$ from $A_t$ by either a user or an agent, the web page transitions to its updated state, labeled as $S_{t+1}$. We use the notation $S_{t+1} = S_t \wedge a_t$ to represent that by executing action $a_t$ on state $S_t$, the system transitions to the new state $S_{t+1}$.

\textbf{Navigation tasks on UI pages.} Given an initial page $S_0$ and a final state $S^*$, the objective of the UI navigation task is to ensure that $S_T = S^*$, which is achieved by sequentially applying a series of decisions $A = \{a_0, a_1, \cdots, a_{T-1}\}$ to transition from $S_0$ to $S_T$ through the operations $S_0 \wedge a_0 \wedge a_1 \wedge \cdots \wedge a_{T-1}$. For each $t \in \{0, \cdots, T-1\}$, the action $a_t$ is chosen from the set $A_t$, which contains all possible interactive actions available on page $S_t$. We usually call $S^*$ the intention or purpose of the UI navigation task.

Therefore, the aim for a tool or model designed for UI navigation is to develop a decision-making function $\hat{a}_t = f(S^*, S_0)$ that achieves $S^* = S_T = S_0 \wedge \hat{a}_0 \wedge \hat{a}_1 \cdots \wedge \hat{a}_{T-1}$, all within a limited number of steps $T$. Previous research~\cite{rawles2024androidinthewild,humphreys2022data} has also incorporated historical operation trajectories and states into the input for decision-making functions in the study of UI navigation tasks. For these approaches, the decision-making function can be represented as $\hat{a}_t = f(S^*, S_0, \cdots, S_{t-1}, a_0, \cdots, a_{t-1})$. Additionally, for ease of annotation and comprehension, the target state $S^*$ is typically described using a single sentence $p$.


\subsection{Data Collections}

We design an annotation systems for annotators to interact with and record tasks to record the real human's behaviour on tiny apps.
The annotation system establishes a real-time connection with an Android emulator in the backend, while the annotators interact with the frontend which contains a mobile interface and task description. The entire data collection process is as follows:
 \emph{(1)}The backend server synchronizes screenshots from the backend Android device to the frontend interface. \emph{(2)}Annotators act on the interface according to specified tasks, such as clicking buttons, scrolling pages, entering content, and navigating back, etc.  \emph{(3)}The backend server records the actions and synchronizes the current screenshot, operation coordinates, and text to the cloud as a record.  \emph{(4)}The backend server sends the operation instructions of the tagging personnel to the Android virtual machine and actually performs the corresponding operations. \emph{(5)}After the emulator performs the actions, it updates the screenshot to the frontend.


\subsection{Data Organization Methods}

Our dataset is composed of 49,023 operational traces across diverse mini-programs within super apps. It encompasses both single-step and multi-step traces, spanning 27 sectors including catering, retail, healthcare, and government services, and extends to over 20,000 distinct tiny-apps and urls. For each operation trace, we will provide the corresponding operation purpose $p$, an indicator of whether the purpose was achieved, and a series of operation steps. At the same time, for each operation step, we will provide page screenshots, page layout analysis results and corresponding actions.

\noindent\textbf{Operation Purpose $p$.} The purpose $p$ is succinctly described in a single sentence, typically suggesting a corresponding page state $S^*$. For instance, the intent to "rent an iPhone 15" suggests that $S^*$ would be the order or payment page for the iPhone 15 within the tiny-app. Our dataset encompasses over 10,000 such action intents. Moreover, for more complex purposes, we also detail the sub-purposes associated with different steps.

\noindent\textbf{Operation Step.} Each operational step is defined by the current state information including a page screenshot, page layout analysis results and the action performed. Executable actions fall into several categories: CLICK, SWIPE, WAIT, INPUT, and STOP. The CLICK action specifies the coordinates where the click occurred, while the SWIPE action contains a start point and an end point to represent the sliding between the two points. 

To gain a deeper insight into our dataset's composition, we have visualized one of its trajectories in Figure~\ref{fig:label-num-trace}. Additionally, Figure~\ref{fig:screenshot} illustrates the data associated with each step of the trajectory alongside the data production pipeline.



\begin{figure*}[tb]
\begin{minipage}{0.32\linewidth}
    \vspace{3pt}
    \centerline{\includegraphics[width=\textwidth]{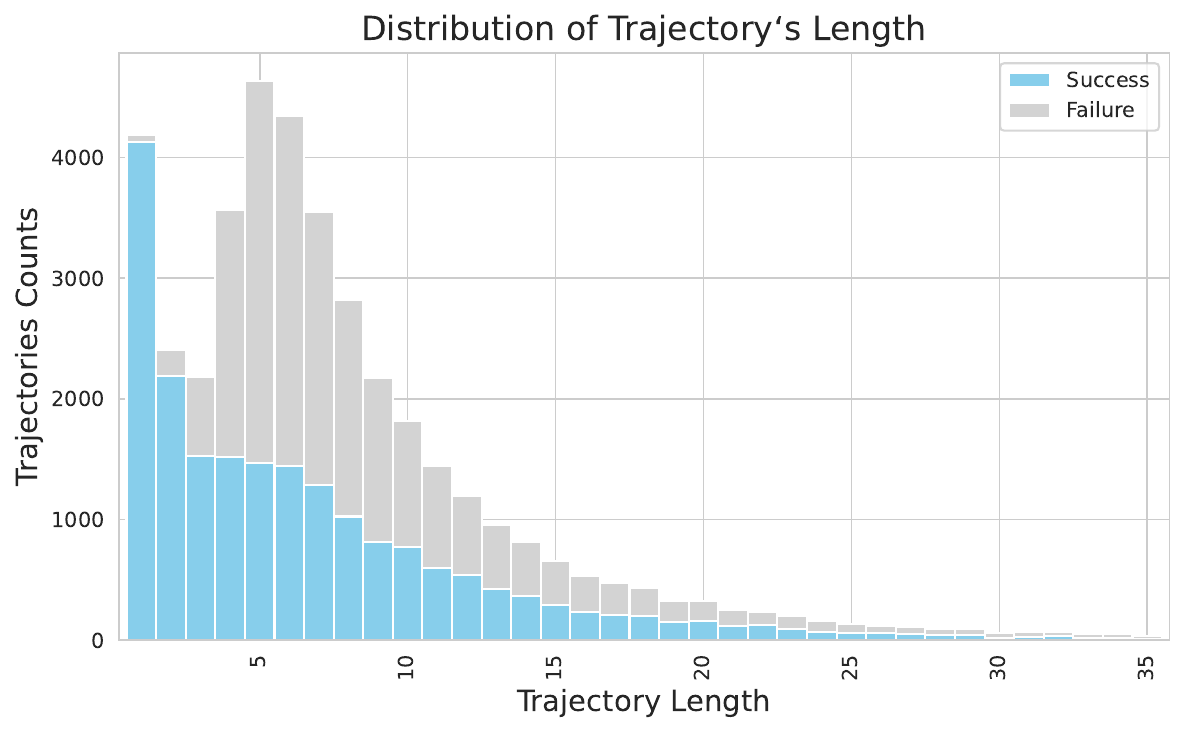}}
    \centerline{(a)}
\end{minipage}
\begin{minipage}{0.32\linewidth}
    \vspace{3pt}
    \centerline{\includegraphics[width
=\textwidth]{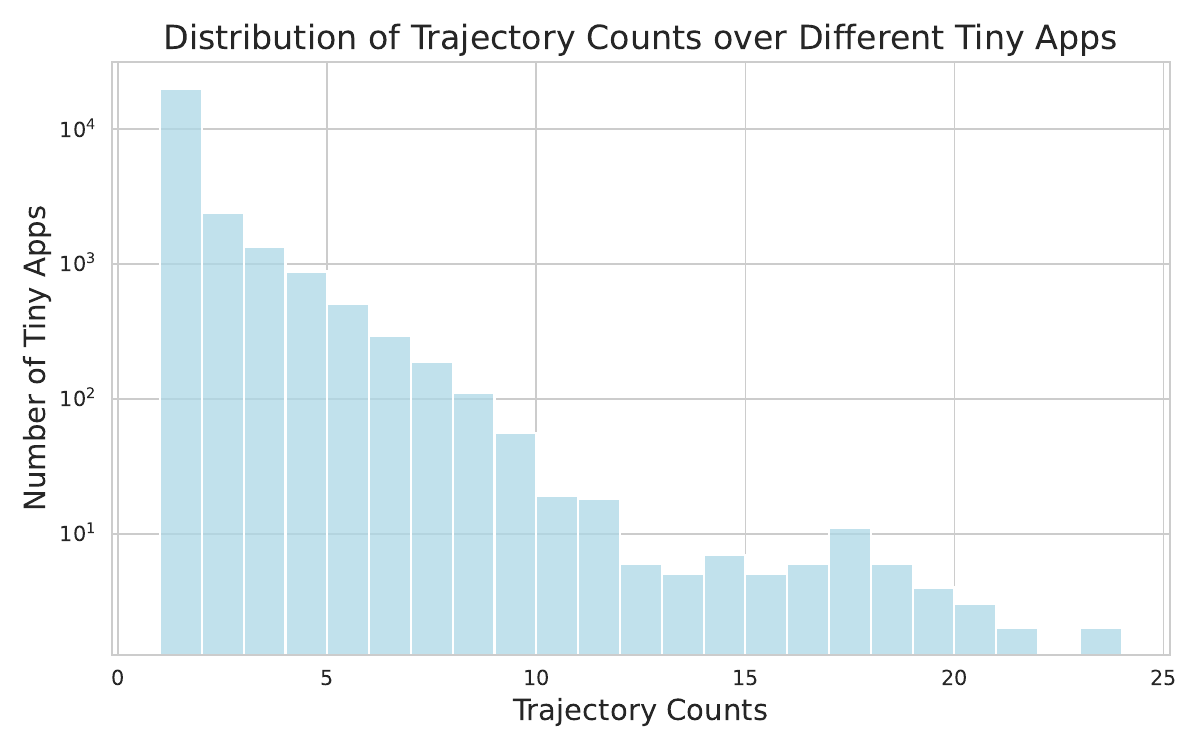}}
    \centerline{(b)}
\end{minipage}
\begin{minipage}{0.32\linewidth}
    \vspace{3pt}
    \centerline{\includegraphics[width=\textwidth]{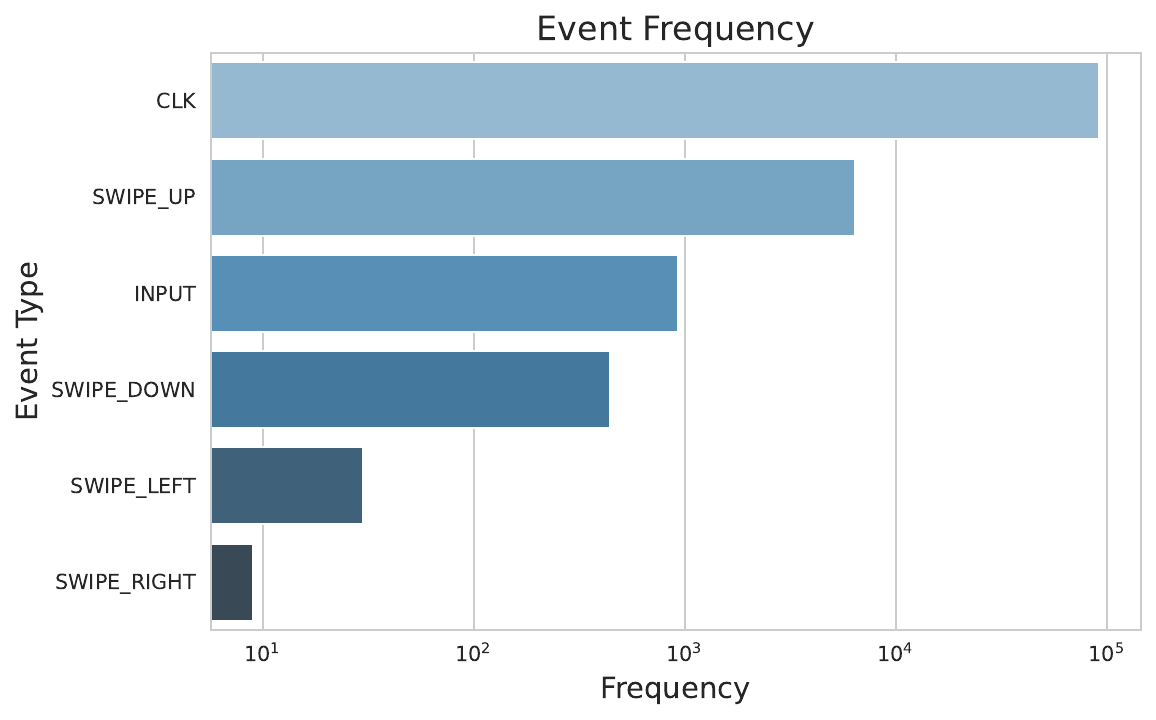}}
    \centerline{(c)}
\end{minipage}
\begin{minipage}{\linewidth}
    \vspace{3pt}
    \centerline{\includegraphics[width=\textwidth]{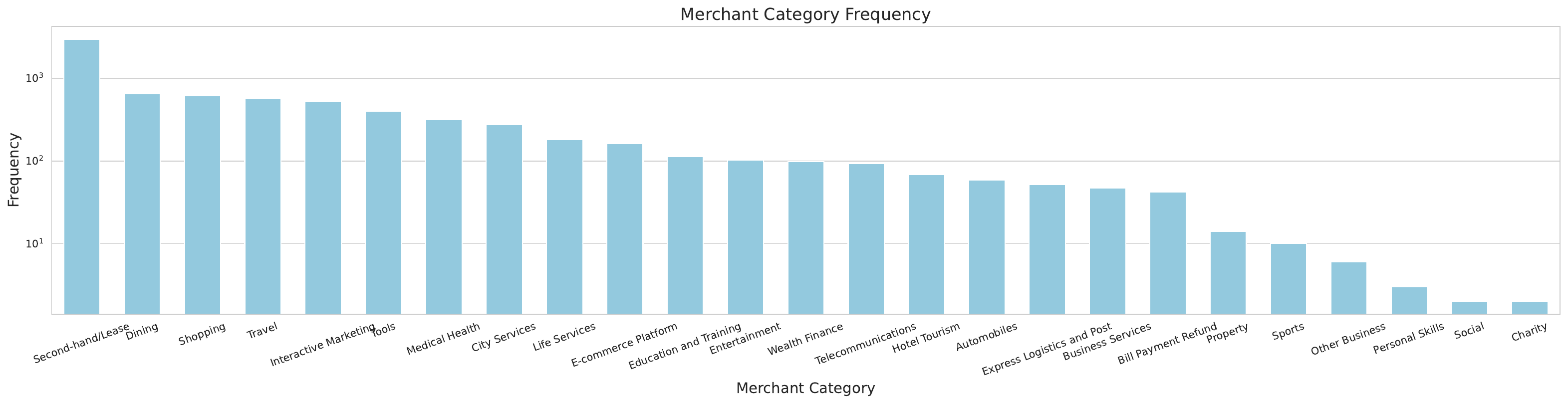}}
    \centerline{(d)}
\end{minipage}

\caption{We illustrate key characteristics of our dataset. (a) provides a overview of trajectories distribution. Most trajectories ranges less than 10 steps, longer trajectories tend to have lower success rate.(b) Our dataset is collected from over 20,000 distinct tiny apps and URLs with very low repetition and high variability. (c) displays the distribution of event type at each individual step, which gathered over 14,0000 single step operations. (4) all tiny apps can be classified into 25 different merchant categories, ranging from release industry to tool, indicating diversity in operation.}
\label{fig:data description from differenct aspect}
\vspace{-4mm}
\end{figure*}

\subsection{Analysis of our Datasets}

\noindent\textbf{More divergence.} We contend that our dataset exhibits greater diversity compared to existing datasets in the UI navigation domain. The data we have gathered originates from mini-apps created by various developers, signifying a range of UI design styles and operational logics. As illustrated in the figure~\ref{fig:data description from differenct aspect}, our dataset encompasses over 20,000 distinct tiny-apps, with the majority appearing in fewer than 5 trajectories within the dataset. This diversity introduces significant challenges for the generalization of models or agents developed or assessed using this dataset. In contrast, most of AitW's data is sourced from Google's first-party applets or those applets developed in close collaboration with Google, with the majority of this data concentrated in specific applications such as Chrome, Android Settings, and Gmail. We also visualize the succes and fail rate at different length of traces, as shown in the figure~\ref{fig:data description from differenct aspect}, which shows that as the step length raise, the success rate get lower, implying that longer steps tend to have a higher failure rate.

\noindent\textbf{Chinese language.} Both MiniWob for web pages and AitW for Android phones focus mostly on English. They don’t have much data for other languages. Since Chinese is one of the world's most used languages on the internet, and because it uses a different writing system from English, it's hard for models trained on English data to work well with Chinese. This means we really need to make a dataset for Chinese UI navigation.

\noindent\textbf{Layout Analysis vs OCR.} In our dataset, we employ a layout analysis algorithm~\cite{DBLP:conf/cvpr/GuXCLMW23} rooted in UI data as an alternative to OCR. As depicted in the figure~\ref{fig:layoutcase}, our layout analysis algorithm outperforms OCR technology by capturing a wider range of UI elements that are likely integral to decision-making processes. When we compare our layout analysis approach to the IconNet methodology used by AitW in the AitW dataset, it is evident that our algorithm identifies a more comprehensive set of elements. Furthermore, our method groups text and icons that are spatially proximate, which better reflects the spatial logic inherent in UI design.



\section{Dataset Evaluation}
\subsection{Evaluation Metrics}

\noindent\textbf{Picture-level Accuracy.}
We employ a key evaluation metric that measures the congruence between the model's decisions in its present state and the corresponding real-world actions observed in each image. This alignment is quantified by posing a binary question. 
For "CLICK" actions, the model must identify the precise click element. 
That is, the model needs to find the element information that matches the expected action (marked by the annotator) from the given layout analysis results and output it.
Meanwhile, the model is considered to have made a correct decision for other action types as long as it accurately predicts the type of action performed. We leverage these criteria to calculate the model's mean accuracy across various images, serving as a gauge of its performance in UI navigation tasks.

\noindent\textbf{Trajectory-level Accuracy.}
In undertaking GUI navigation tasks, each trajectory consists of a sequence of steps. An error in predicting any one of these steps could hinder the successful navigation to the intended destination. Consequently, we consider the model's trajectory-level accuracy as a key performance measure. Successful navigation is achieved only when the model executes the appropriate action for each image within a given trajectory. 



\subsection{Methods}
We employ three evaluation methods to assess the performance of various models on the E-ANT, specifically designed for LLMs/MLLMs. These methods include Zero-shot inference and Fine-tuning. In addition, we trained an XYLayoutLM model~\cite{DBLP:conf/cvpr/GuMWLWG022} trained using the behavioral cloning method, which is a baseline for non-generative methods.

\begin{table*}[t]
\centering
\caption{Benchmarks on E-ANT cover LLM, MLLM and non-generative models}\label{tab:tab2}
\label{tab:baseline_res}
\begin{tabular}{|l|l|r|r|}
\toprule
model type & training strategy & Picture-level acc  & Trajectory-level acc \\
\midrule 
GPT-3.5-16K & Zero-shot & 23.5\% & 1.9\% \\
\midrule 
LLaVA-v1.5-7B & Zero-shot & 12.8\% & 0.4\% \\
LLaVA-NeXT-7B & Zero-shot & 19.9\% & 0.8\% \\
\midrule 
Qwen-72B & Zero-shot& 28.6\% & 2.1\% \\
Qwen1.5-14B & Zero-shot& 18.7\% & 0.6\% \\
\midrule 
\textbf{XYLayoutLM} & \textbf{Finetune(Behavioral Cloning)} & \textbf{66.8\%} & \textbf{11.1\%} \\ 
\midrule 
\textbf{LLaVA-v1.5-7B} & \textbf{Finetune}& \textbf{47.3\%} & \textbf{3.6\%} \\
\midrule 
\textbf{LLaVA-v1.5-7B} & \textbf{Finetune with data augmentation }& \textbf{51.6\%} & \textbf{4.0\%} \\
 \bottomrule
 
\end{tabular}
\vspace{-4mm}
\end{table*}

\noindent\textbf{Zero-shot inference for LLM/MLLMs.}
In inference using Language Models (LLMs), we systematically analyze the layout information presented by each image and provide it to LLMs. 
This information is captured within a structured element bar, which adheres to a standardized format: $\{ 'id': <\cdot>, 'cate': <\cdot>, 'text': <\cdot>, 'box': <\cdot> \}$. Once formatted, this data is then fed into the LLM, which is instructed to output responses that conform to a predetermined structure, given by the template: $\{ 'thinking': <\cdot>, 'action\_type': <\cdot>, 'button': \{ 'id': <\cdot>, 'cate': <\cdot>, 'text': <\cdot>, 'box': <\cdot> \} \}$. Based on this scheme, the LLM selects an appropriate action; however, if the `$action\_type$' is not a `$click$', the details of the `button' need not be furnished. 
Meanwhile, 
We supply MLLMs with both layout parsing text prompts and original image embeddings.

\noindent\textbf{Fine-tuning for MLLMs.}
In addition to the setting without retraining, we also focus on the performance of multi-modal large models after fine-tuning instructions for certain UI navigation task data. We referred to the training method provided by Auto-UI~\cite{zhan2023you}, used pictures as model input, and then directly organized the decision results into text as the object that the model needed to learn. In organizing training data for fine-tuning, we categorize it into two types: general domain data, sourced from LLaVA~\cite{liu2024visual,liu2023improved}, and data specifically aimed at GUI (Graphical User Interface) navigation decisions. All LLaVA-based fine-tuning results are obtained by training one epoch using the mixed data on the official checkpoint released by the company. The training settings refer to the standard settings.

\noindent\textbf{Fine-tuning for XYLayoutLM.}
We formulate the decision task as a combination of a NER(Named Entity Recognition) task and a sentence-level classification task using a multimodal model, and train these two tasks on the XYLayoutLM model~\cite{DBLP:conf/cvpr/GuMWLWG022}, which is an improvement of the LayoutLM family of models. Most of the model's raw settings are kept except each detected element is treated as an delicated word with text content like 'icon: <ocr text>' and its corresponding position. The sentence-level classification task is used to learn the expected action type on the current page, which will make judgments among types such as CLICK, WAIT, SWIPE, INPUT, SUCCESS and FAIL. For CLICK or INPUT action, the model is also required to highlight the token corresponding to the element expected to be tapped on. This is described as an NER task, that is a binary classification task in the token dimension to decide which elemented to be chosen. As for the INPUT part, since XYLayoutLM is not a generative model, so we treat a action is correct as long as the model can identify the correct action and token.

\subsection{Experiments Results on E-ANT}
We randomly selected 1000 trajectories as our testset (about 5\%), and for fine-tuning experiments, we selected 4000 trajectories (about 20\%) that are not included in the testset as training data. We present our experimental results in the table~\ref{tab:baseline_res}. We can notice that for zero-shot inference, although most models can correctly infer the decision under a single step on some pictures, they perform poorly in terms of accuracy in the trajectory-level. At the same time, for the same zero-shot inference, LLaVA did not show a higher level of competitiveness than stronger LLMs such as GPT3.5 and Qwen72B, even though it added image modal input. This may be because under the premise of carrying layout parsing text input, increasing the basic ability of the text base itself is more important for decision-making than adding a modal input. In addition, the fine-tuned LLaVA can show higher accuracy than the un-fine-tuned version. 
Combining the understanding of data through GUI with decision-making data effectively enhances its accuracy.

\section{Data Augmentation Method for E-ANT}
\subsection{Motivation}

Typically, human users navigate through GUI navigation tasks in a two-step approach: initially by comprehending the contents of the page (page understanding), and subsequently determining the necessary actions (action decision-making). This indicates that a profound understanding of the page significantly enhances the model’s ability to connect the input image with specific actions.

Generally speaking, if you want the model to obtain a good understanding of UI pages, you need to rely on additional training data, including page understanding data sets such as UI-Bert and RICO. 
However, a challenge arises due to the stylistic divergence between the UI designs in these datasets and those encountered in current UI navigation tasks. This discrepancy may lead to a disjointed integration of the two critical phases of understanding and action.

\begin{figure*}[ht]
	\begin{center}
  \includegraphics[width=0.95\textwidth]{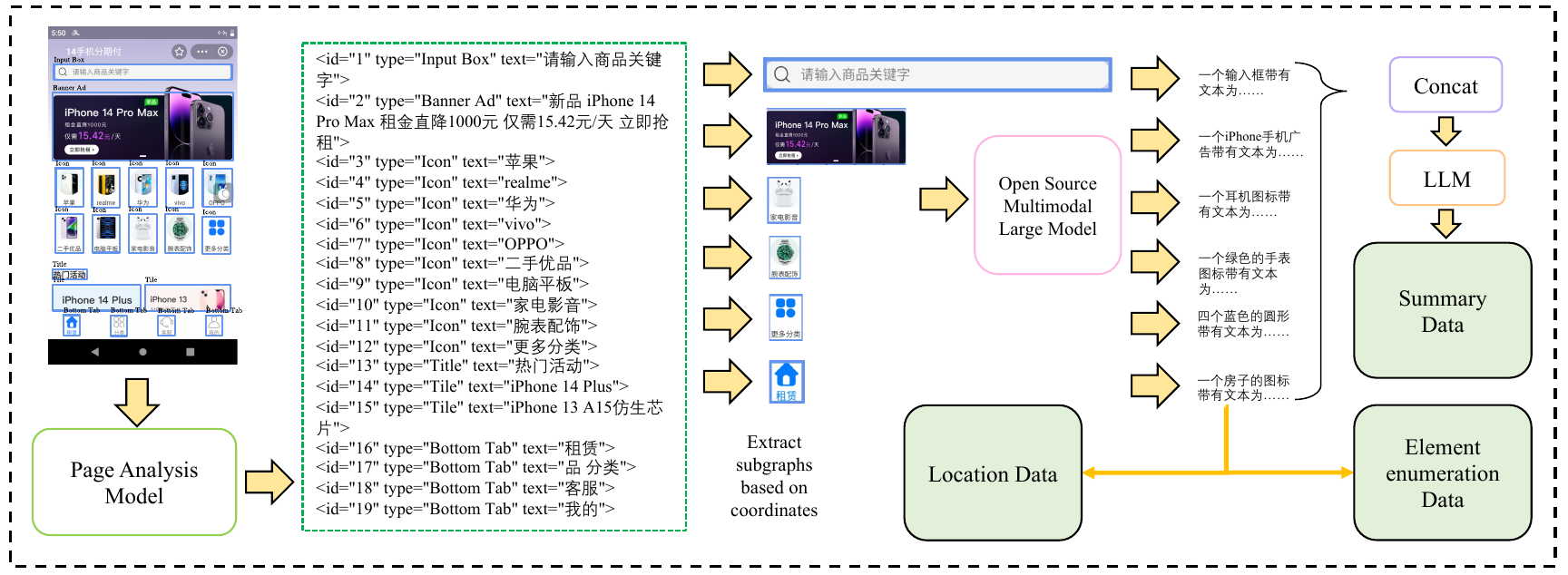} \\
	\end{center}
	\caption{Pipeline for executing data augmentation methods on E-ANT.} \label{data_gen_ana}
	\vspace{-4mm}
\end{figure*}
In fact, the image trajectory data utilized for GUI navigation tasks is 
valuable for understanding image pages. However, it lacks the necessary annotation information for each element within the image. Simultaneously, employing annotators to add further annotations to all images and their respective elements would entail a significant investment in terms of both manpower and financial resources. Another approach involves leveraging the existing MLLM for auxiliary annotation. However, we've observed that the current MLLM struggles to fully and accurately identify elements within GUI images. This challenge arises from the characteristic of the MLLM's training data, which primarily consist of regular images and not specialized GUI content like app screenshots.
To enhance the image trajectory data from GUI navigation tasks into high-quality image page understanding data without incurring additional costs, this paper introduces a bootstrap data augmentation method. This method leverages the existing layout parsing model and the MLLMs to expand data.

\subsection{Overview and data workflow}

In this subsection, we describe our approach to training our multimodal UI navigation model using UI navigation data. Our methodology is structured around two critical processes: generating page understanding data and creating chained decision-making data. The goal of the first process is to enrich the model's comprehension of webpage content, while the second process is designed to improve the model's ability to link navigation objectives with ultimate decision-making actions.

\noindent\textbf{Page understanding data generation.} For processing an GUI page, we commence by deploying the GUI layout parsing model to identify and segment the page elements into multiple sub-images. Each sub-element image is then fed into an advanced multi-modal large model (such as LLaVA, MiniGPT4, or Blip) that does not specialize in UI navigation data fine-tuning, requiring the model to independently generate an outline. Through this approach, we capture both the coordinates of each element (via layout analysis) and their descriptive information (via the multi-modal model).
Subsequently, we synthesize three distinct types of data for a comprehensive understanding of the page: element positioning data, page element enumeration data, and a page summary. Examples of these data types are illustrated in accompanying figures. To create the element positioning data, we simply pair the coordinates with their respective descriptions for each element. The page element enumeration data is produced by aggregating these pairs across all page elements into a cohesive paragraph. Finally, for the page summary data, we compile the descriptions of all elements and submit them to the LLM to generate a succinct summary.

\noindent\textbf{Chained decision-making data generation.} Indeed, for numerous pages and navigational goals, the link between the intended navigation and the specific actions required on a given page is not always intuitive. Even humans must meticulously review the page's content before deciding on their navigational approach. 
To improve the model’s ability to establish this connection, we used multi-round conversation data in the training data, requiring the model to first answer questions related to the understanding of the page and then make corresponding action decisions.




\section{Conclusion}
Navigation plays an important rule in people's daily life, yet we find there is a lack of a comprehensive and well-designed data benchmark. Moreover, existing benchmarks are predominantly in English, with poor box quality and limited availability for Chinese. To address these issues, we have introduced a new benchmark with a large-scale dataset and several distinct features, including more divergence and good layout baselines.
For now we collect over 40k high quality trajectories performed and corrected by human annotators, which will fill the gaps of navigation data on Mobile UI.

\section{Limitation}

The E-ANT dataset is annotated by the annotators on the computer page through ADB interacting with the Android virtual machine in the background, which means that there is still a gap between our environment and the real Android mobile phone, and it cannot have more flexible operations like using the Android system directly on the mobile phone.

In addition, although E-ANT is the first large-scale Chinese GUI navigation dataset with a large amount of data, due to the existence of a large number of heterogeneous mobile devices and Chinese applications with different resolutions and GUI styles, there is still a need to further improve the scope and quality of the data to help build a more effective GUI navigation intelligent agent.

Finally, various high-performance LLM/MLLMs are being released by different researchers. Due to limited resources (computing resources and open model checkpoint resources), we, as data publishers, cannot traverse and evaluate all models on the market using E-ANT data. However, we will work with the community to improve such evaluations as much as possible and continuously iterate and optimize the training and testing pipelines.

\bibliography{custom}
\appendix



\end{document}